\documentclass{tlp}

\title[On Finitely Recursive Programs]
        {On Finitely Recursive Programs\footnote{This paper extends and refines \cite{DBLP:conf/iclp/BaseliceBC07}}}
	
  \author[S. Baselice, P.A. Bonatti, G. Criscuolo]
         {Sabrina Baselice, Piero A. Bonatti, Giovanni Criscuolo\\
         Universit\`a di Napoli ``Federico II'', Italy\\
         }
  
 \submitted {7 April 2008}
 \revised {30 December 2008}
 \accepted{16 January 2009}

  \usepackage{amsmath,amssymb}
  \usepackage{times}
  \usepackage{latexsym}
  \usepackage{graphicx}
  \usepackage{xspace}
  \usepackage{url}
  \usepackage{algorithm}
  \usepackage[noend]{algorithmic}

  \newcommand{\hide}[1]{}
  \newcommand{\naf}{\ensuremath{\mathop{\tt{not}}}\xspace}
  \newcommand{\tup}[1]{\ensuremath{\langle #1 \rangle}\xspace}

  \newcommand{\ground}{\mathsf{Ground}}
  \newcommand{\atom}{\mathit{atom}}
  
  \newcommand{\lm}{\mathsf{lm}}

  \newcommand{\SM}{\ensuremath{\mathit{SM}}\xspace}
  
  \newcommand{\CounterSupp}{\ensuremath{\mathsf{CounterSupp}}\xspace}
  \newcommand{\true}{\ensuremath{\mathsf{TRUE}}\xspace}
  \newcommand{\false}{\ensuremath{\mathsf{FALSE}}\xspace}

\newtheorem{theorem}{Theorem}[section]
\newtheorem{lemma}[theorem]{Lemma}
\newtheorem{corollary}[theorem]{Corollary}
\newtheorem{proposition}[theorem]{Proposition}
\newtheorem{definition}[theorem]{Definition}
\newtheorem{example}[theorem]{Example}

{
    \renewcommand{\theenumi}{\alph{enumi}}
    
    \begin{enumerate}
}%
{
    \end{enumerate}
}

\newenvironment{enumroman}%
{
    \renewcommand{\theenumi}{\roman{enumi}}
    
    \begin{enumerate}
}%
{
    \end{enumerate}
}

\newtheorem{exa}[theorem]{Example} 

\begin{document}

\maketitle

\begin{abstract}
  Disjunctive \emph{finitary programs} are a class of logic programs admitting
  function symbols and hence infinite domains.  They have very good
  computational properties, for example ground queries are decidable while in the
  general case the stable model semantics is $\Pi^1_1$-hard. In this
  paper we prove that a larger class of programs, called
  \emph{finitely recursive programs}, preserves most of the good
  properties of finitary programs under the stable model semantics,
  namely: (i) finitely recursive programs enjoy a compactness
  property; (ii) inconsistency checking and skeptical reasoning are
  semidecidable; (iii) skeptical resolution is complete for normal finitely recursive programs. Moreover, we
  show how to check inconsistency and answer skeptical queries using
  finite subsets of the ground program instantiation.
  We achieve this by extending the splitting sequence theorem by Lifschitz
  and Turner: We prove that if the input
  program $P$ is finitely recursive, then the partial stable
  models determined by any smooth splitting $\omega$-sequence converge
  to a stable model of $P$.
\end{abstract}

\section{Introduction}

Answer Set Programming (ASP) \cite{DBLP:journals/corr/cs-LO-9809032,DBLP:journals/amai/Niemela99} is one of the most interesting achievements in the area of Logic Programming and Nonmonotonic Reasoning.  It is a declarative problem solving paradigm, mainly centered around some well-engineered implementations of the stable model semantics of logic programs \cite{GL88,gelfond91classical}, such as  \textsc{Smodels} \cite{smodels} and DLV \cite{DLV}. 

The most popular ASP languages are extensions of Datalog, namely, function-free, possibly disjunctive logic programs with negation as failure.  The lack of function symbols has several drawbacks, related to expressiveness and encoding style  \cite{DBLP:journals/ai/Bonatti04}.  In order to overcome such limitations and reduce the memory requirements of current implementations, a class of logic programs called \emph{finitary programs} has been introduced  \cite{DBLP:journals/ai/Bonatti04}.

In finitary programs function symbols (hence infinite domains) and recursion are allowed.  However, recursion is restricted by requiring each ground atom to depend on finitely many ground atoms; such programs are called \emph{finitely recursive}.  Moreover, only finitely many ground atoms must occur in \emph{odd-cycles}---that is, cycles of recursive calls involving an odd number of negative subgoals---which means that there should be only finitely many potential sources of inconsistencies.  These two restrictions bring a number of nice semantical and computational properties  \cite{DBLP:journals/ai/Bonatti04}.  In general, function symbols make the stable model semantics highly undecidable \cite{MR01expressibility}.  On the contrary,  if the given program is finitary, then consistency checking, ground credulous queries, and ground skeptical queries are decidable.  Nonground queries were proved to be r.e.-complete.  Moreover, a form of compactness holds: an inconsistent finitary program has always a finite \emph{unstable kernel}, i.e.\ a finite subset of the 
ground instantiation of the program with no stable models.  All of these properties are quite unusual for a nonmonotonic logic.

As function symbols are being integrated in state-of-the-art reasoners such as DLV \cite{DBLP:conf/iclp/CalimeriCIL08},
it is interesting to extend these good properties to larger program
classes.  This goal requires a better understanding of the role of each restriction in the definition of finitary programs.    It has already been noted \cite{DBLP:journals/ai/Bonatti04}
that by dropping the first condition (i.e., if the program is not finitely recursive) one obtains a superclass of stratified programs, whose
complexity is then far beyond computability.  In the same paper, it is argued that the second restriction (on odd-cycles) is needed for the decidability of ground queries.  However, if a program is only finitely recursive (and infinitely many odd-cycles are allowed), then the results of \cite{DBLP:journals/ai/Bonatti04} do not characterize the exact complexity of reasoning and say nothing about compactness, nor about the completeness of the skeptical resolution calculus \cite{DBLP:journals/jar/Bonatti01}.

In this paper we extend and refine those results, and prove that
several important properties of finitary programs carry over to all
disjunctive finitely recursive programs.  We prove that for all such
programs the compactness property still holds, and that inconsistency
checking and skeptical reasoning are semidecidable.  Moreover, we
extend the completeness of skeptical resolution
\cite{DBLP:journals/jar/Bonatti01,DBLP:journals/ai/Bonatti04} to all
normal finitely recursive programs.  Our results clarify the role that
each of the two restrictions defining normal finitary programs has in
ensuring their properties.

In order to prove these results we use program splittings
\cite{lifschitz94splitting}, but the focus is shifted from splitting
sequences (whose elements are sublanguages) to the corresponding
sequences of subprograms, that enjoy more invariant properties and may
be regarded as a sort of normal form for splitting sequences.  For
this purpose we introduce the notion of \emph{module sequence}.  It
turns out that disjunctive finitely recursive programs are exactly
those disjunctive programs whose module sequences consist of finite
elements. Moreover a disjunctive finitely recursive program $P$ has a
stable model whenever each element $P_i$ of the sequence has a stable
model, a condition which is not valid in general for all disjunctive
programs with negation.  This result provides an iterative
characterization of the stable models of $P$.  Module sequences and
this theorem constitute a powerful formal tool that may facilitate the
proof of new consistency results, and provide a uniform framework for
comparing different approaches to decidable reasoning with infinite
domains.

The paper is organized as follows. The next section is devoted to
preliminaries.  In Section~\ref{SplitSeq}, we define
module sequences and study their properties.  In
Section~\ref{sec:compactness}, we prove that every finitely recursive program with a consistent module sequence is consistent, and use this result to
extend the compactness property of finitary programs to all finitely
recursive programs.  Complexity results and two simple sound and
complete algorithms for inconsistency checking and skeptical reasoning
can be found in Section~\ref{sec:complexity}.  Then, for a better,
goal-directed calculus, the completeness theorem for skeptical
resolution is extended to all finitely recursive programs in
Section~\ref{sk-resolution}.  Section~\ref{sec:finitary} relates
finitely recursive programs and our iterative approach to previous
approaches to decidable reasoning with infinite stable models, and makes a first step towards a unified picture based on our framework.
Finally, Section~\ref{conclusions} concludes the paper with a summary
and a brief discussion of our results, as well as some interesting
directions for future research.

\section{Preliminaries}		\label{preliminaries}

We assume the reader to be familiar with the classical theory of logic 
programming \cite{DBLP:books/sp/Lloyd84}.

\emph{Disjunctive logic programs} are sets of (disjunctive) rules
  \begin{equation*}
     A_1 \vee A_2 \vee ... \vee A_m \leftarrow L_1, ..., L_n \quad
  \quad (m> 0, n\geq 0) ,
  \end{equation*}
  where each $A_j$ ($j=1, ..., m$) is a logical
  atom and each $L_i$ ($i=1, ..., n$) is a \emph{literal}, that is,
  either a logical atom $A$ or a negated atom $\naf A$.

If $r$ is a rule with the above structure, then let $head(r)=\{A_1,$
$A_2, ...,$ $A_m\}$ and $body(r)=\{L_1, ...,$ $L_n\}$.  Moreover, let
$body^+(r)$ (respectively $body^-(r)$) be the set of all atoms $A$
such that $A$ (respectively $\naf A$) belongs to $body(r)$.

\emph{Normal logic programs} are disjunctive logic programs whose
rules $r$ have one atom in their head, that is, $|head(r)|=1$.

The ground instantiation of a program $P$ is denoted by $\ground(P)$,
and the set of atoms occurring in $\ground(P)$ is denoted by
$\atom(P)$.  Similarly, $\atom(r)$ denotes the set of atoms occurring
in a ground rule $r$.

A Herbrand model $M$ of $P$ is a \emph{stable model} of $P$ iff $M\in
\lm(P^M)$, where $\lm(X)$ denotes the set of least models of a
positive (possibly disjunctive) program $X$, and $P^M$ is the
\emph{Gelfond-Lifschitz transformation} \cite{GL88,gelfond91classical} of $P$, obtained from
$\ground(P)$ by 
\begin{enumroman}
\item removing all rules $r$ such that $body^-(r) \cap M \neq
  \emptyset$, and
\item  removing all negative literals from the body
  of the remaining rules.
\end{enumroman}

Disjunctive and normal programs may have one, none, or multiple stable
models.  We say that a program is \emph{consistent} if it has at least
one stable model; otherwise the program is \emph{inconsistent}.  A
\emph{skeptical} consequence of a program $P$ is any closed first order formula satisfied
by all the stable models of $P$.  A \emph{credulous} consequence of
$P$ is any closed first order formula satisfied by at least one stable model of $P$.

The \emph{dependency graph of a program $P$} is a labelled directed
graph, denoted by $DG(P)$, whose vertices are the ground atoms of
$P$'s language. Moreover,
\begin{enumroman}
\item there exists an edge labelled `+' (called positive edge) from
  $A$ to $B$ iff for some rule $r\in \ground(P)$, $A\in head(r)$ and
  $B\in body(r)$;
\item there exists an edge labelled `-' (called negative edge) from
  $A$ to $B$ iff for some rule $r\in \ground(P)$, $A\in head(r)$ and
  $\naf B\in body(r)$;
\item there exists an unlabelled edge from $A$ to $B$ iff for some rule $r\in
  \ground(P)$, $A\in head(r)$ and $B\in head(r)$.
\end{enumroman}

An atom $A$ \emph{depends positively} (respectively \emph{negatively}) on $B$ 
if there is a directed path from $A$ to $B$ in the dependency graph with an 
even (respectively odd) number of negative edges. Moreover, each atom depends 
positively on itself. $A$ \emph{depends} on $B$ if $A$ depends
positively or negatively on $B$.

  An \emph{odd-cycle} is a cycle in the dependency graph with an odd number 
  of negative edges. A ground atom is \emph{odd-cyclic} if it occurs 
  in an odd-cycle.
%
Note that there exists an odd-cycle iff some ground atom $A$ depends
negatively on itself.

The class of programs on which this paper is focussed can now be
defined very concisely.

\begin{definition}
\label{def3}
  A disjunctive program $P$ is \emph{finitely recursive} iff each
  ground atom $A$ depends on finitely many ground atoms in
  $DG(P)$\footnote{
    This definition differs from the one adopted in
  \cite{DBLP:conf/iclp/Bonatti02} because it is based on a different
  notion of dependency.  Here the dependency graph contains
  edges between atoms occurring in the same head, while in
  \cite{DBLP:conf/iclp/Bonatti02} such dependencies are dealt with in
  a third condition in the definition of finitary programs.  Further
  comparison with \cite{DBLP:conf/iclp/Bonatti02} can be found in
  Section~\ref{sec:finitary}.   }.
\end{definition}

For example, most standard list manipulation programs 
($\mathtt{member, append,\ remove}$ etc.) are finitely recursive.  The reader 
can find numerous examples of finitely recursive programs in 
\cite{DBLP:journals/ai/Bonatti04}.  
In general, checking whether a program is finitely recursive is undecidable \cite{DBLP:journals/ai/Bonatti04}.  
However, in \cite{BoLPNMR01,DBLP:journals/ai/Bonatti04} a large decidable subclass has been implicitly characterized via static analysis techniques. 
Another expressive, decidable class of finitely recursive programs can be found in \cite{DBLP:conf/lpar/SimkusE07}.

We will also mention frequently an important subclass of finitely
recursive programs:

\begin{definition}[Finitary programs]
\label{def4}
  We say that a disjunctive program $P$ is \emph{finitary} if the following conditions hold:
  \begin{enumerate}
  \item \label{def4prop1} $P$ is finitely recursive.
  \item \label{def4prop2} There are finitely many odd-cyclic atoms in the dependency graph 
  $DG(P)$.
  \end{enumerate}
\end{definition}

Finitary programs have very good computational properties (for example
ground inferences are decidable).  Many interesting programs, however,
are finitely recursive but not finitary, due to integrity constraints
that apply to infinitely many individuals.

\begin{example}
Typical programs for reasoning about actions and change
are finitary.
Fig.\ 4 of \cite{DBLP:journals/ai/Bonatti04} illustrates one of them,
modelling a blocks world. That program defines---among others---two
predicates $\mathtt{holds}(\mathit{fluent,time})$ and
$\mathtt{do}(\mathit{action,time})$.  The simplest way to add a
constraint that forbids any parallel execution of two incompatible
actions $a_1$ and $a_2$ is including a rule $$f \leftarrow \naf f,
\mathtt{do}(a_1,T), \mathtt{do}(a_2,T)$$ in that program, where $f$ is
a fresh propositional symbol (often such rules are equivalently
expressed as \emph{denials} like $\leftarrow \mathtt{do}(a_1,T), \mathtt{do}(a_2,T)$).  This program is not finitary (because $f$ depends on infinitely many atoms since $T$ has an infinite range of values) but it can be reformulated as a finitely recursive program by replacing the above rule with
\begin{equation*}
	f(T) \leftarrow \naf f(T), \mathtt{do}(a_1,T), \mathtt{do}(a_2,T) \,.
\end{equation*}
Note that the new program is finitely recursive but not finitary, because the new rule introduces infinitely many odd cycles (one for each instance of $f(T)$). 
\end{example}

Our results on finitely recursive programs depend on the \emph{splitting
theorem} that allows to construct stable models in stages.  In turn,
this theorem is based on the notion of \emph{splitting set}.

\begin{definition}[Splitting set and bottom program \cite{Baral03,lifschitz94splitting}]
  A \emph{splitting set} of a disjunctive logic program $P$ is any set $U$ of ground
  atoms such that, for all rules $r\in \ground(P)$, if $head(r)\cap U
  \neq \emptyset$ then $\atom(r)\subseteq U$. If $U$ is a splitting
  set for $P$, we also say that $U$ splits $P$. The set of rules $r\in
  \ground(P)$ such that $head(r)\cap U\neq\emptyset$ is called the
  \emph{bottom} of $P$ relative to the splitting set $U$ and is
  denoted by $bot_U(P)$.
  The subprogram $\ground(P)\setminus bot_U(P)$ is called the 
  \emph{top} of $P$ relative to $U$.
\end{definition}
The bottom program characterizes the restriction of the stable models of $P$ to the
language determined by the splitting set. The top program determines
the rest of each stable model; for this purpose it should be partially
evaluated with respect to\ the stable models of the bottom.
\begin{definition}[Partial evaluation \cite{Baral03,lifschitz94splitting}]
  The \emph{partial evaluation} of a disjunctive logic program $P$ with splitting set $U$ 
  with respect to a set of ground atoms $X$ is the program $e_U(\ground(P),X)$ defined as follows:

\begin{align*}
    e_U(\ground(P),X) = &\{ r'\ \mid \text{ there exists } r\in \ground(P) \text{ s.t. } (body^+(r)\cap U)\subseteq X \\
    &\text{and } (body^-(r)\cap U) \cap X= \emptyset, \text{ and } head(r')=head(r), \\
    & body^+(r')=body^+(r)\setminus U, body^-(r')=body^-(r)\setminus U\, \} \,.
  \end{align*}

\end{definition}
We are finally ready to formulate the splitting theorem (and hence the
modular construction of stable models based on the top and bottom
programs) in formal terms.  
\begin{theorem}[Splitting theorem \cite{lifschitz94splitting}]
  Let $U$ be a splitting set for a disjunctive logic program $P$. An interpretation $M$
  is a stable model of $P$ iff $M=I \cup J$, where 
  \begin{enumerate}
    \item $I$ is a stable model of $bot_U(P)$, and
    \item $J$ is a stable model of $e_U(\ground(P)\setminus bot_U(P),I)$. 
  \end{enumerate}
\end{theorem}

The splitting theorem has been extended to \emph{transfinite sequences} in  
\cite{lifschitz94splitting}.
  A (transfinite) sequence is a family whose index set is an 
  initial segment of ordinals, $\{\alpha\ :\ \alpha < \mu\}$.
  The ordinal $\mu$ is the \emph{length} of the sequence.

  A sequence $\langle U_\alpha \rangle_{\alpha<\mu}$ of sets is \emph{monotone}
  if $U_\alpha \subseteq U_\beta$ whenever $\alpha < \beta$, and 
  \emph{continuous} if, for each limit ordinal $\alpha < \mu$,
  $U_\alpha = \bigcup_{\nu<\alpha} U_\nu$.

\begin{definition}[Lifschitz-Turner, \cite{lifschitz94splitting}]
  A \emph{splitting sequence} for a disjunctive program $P$ is a mo\-notone,
  continuous sequence $\langle U_\alpha \rangle_{\alpha<\mu}$
  of splitting sets for $P$ such that $\bigcup_{\alpha<\mu} U_\alpha= \atom(\ground(P))$\,.
\end{definition}

Lifschitz and Turner generalize the splitting theorem to splitting
sequences.
\begin{theorem}[Splitting sequence theorem \cite{lifschitz94splitting}]
  \label{thm:splitting-seq}
  Let $P$ be a disjunctive program.\footnote{
    The splitting sequence theorem holds for disjunctive logic programs extended with so-called \emph{strong negation} that, however, is essentially syntactic sugar.  Therefore, for the sake of simplicity, we ignore it here.
  }
  $M$ is a stable model of $P$ iff
  there exists a splitting sequence $\langle U_\alpha
  \rangle_{\alpha<\mu}$ such that
  \begin{enumerate}
  \item $M_0$ is a stable model of $bot_{U_0}(P)$, 
  \item for all successor ordinals $\alpha<\mu$, $M_{\alpha}$ is a
    stable model of $e_{U_{\alpha -1}}(bot_{U_\alpha}(P)\setminus
    bot_{U_{\alpha-1}}(P), \bigcup_{\beta<\alpha}M_\beta)$, 
  \item for all limit ordinals $\lambda<\mu$, $M_\lambda=\emptyset$,
  \item $M=\bigcup_{\alpha<\mu} U_\alpha$.
  \end{enumerate}
\end{theorem}

	\section{Module sequences and a normal form for splitting sequences}
	\label{SplitSeq}

In this section we replace the sequences of program slices
$bot_{U_\alpha}(P)\setminus bot_{U_{\alpha-1}}(P)$ adopted by
Lifschitz and Turner with slightly different and simpler program
module sequences. Then we prove some properties of module sequences
that will be useful in proving our main results.
    
\begin{definition}[GH, Module sequence]
\label{defSplitPart}
  Let $P$ be a disjunctive program and let the set of its \emph{ground head atoms} be
  \begin{equation*}
    GH=\{\, p\ \mid\ p\in head(r),\ r\in \ground(P) \,\}.
  \end{equation*}

  The \emph{module sequence $P_1, P_2,$ $...,$ $P_n, ...$ induced by an
  enumeration $p_1, p_2,...,$ $p_n, ...$ of $GH$}
  is defined as follows: 

  \begin{align*}
    P_1 & = \{\, r\in \ground(P)\ \mid\ p_1 \text{ depends on some } A\in head(r) \, \}
\\
    P_{i+1} & = P_i\cup \{ \, r\in \ground(P)\ \mid\ p_{i+1} \text{ depends 
    on some } A\in head(r) \, \} && (i\geq 1).
  \end{align*}
  
\end{definition}

\hide{
\begin{exa}
  Consider the finitely recursive program $P$ of the example~\ref{exa:finitely-rec-prog}:
  \[
  \begin{array}{l}
    p(f(X))\leftarrow p(X), q(X).\\
    q(X)\leftarrow s(X).\\
    u(X)\leftarrow \naf u(X).\\
    z(X)\leftarrow p(X).
  \end{array}
  \]
  The Herbrand Universe of $P$ is $\{a,f(a),f(f(a)),...\}$.
  One enumeration of the $GH$ is
  $e=\{p(f(a)),$ $q(a),$ $u(a),$ $z(a),$ $p(f(f(a))),$ 
  $q(f(a)),$ $u(f(a)),$ $z(f(a)),p(f(f(f(a)))),...\}$.
  A \emph{module sequence for $P$ induced by the enumeration $e$
  of $GH$} is
  \[
  \begin{array}{l}
    P_1=\{p(f(a))\leftarrow p(a),q(a); q(a)\leftarrow s(a)\};\\
    P_2=P_1;\\
    P_3=P_2 \cup \{u(a)\leftarrow \naf u(a)\};\\
    P_4=P_3 \cup \{z(a)\leftarrow p(a)\};\\
    P_5=P_4 \cup \{p(f(f(a)))\leftarrow p(f(a)),q(f(a)); q(f(a))\leftarrow s(f(a))\};\\
    P_6=P_5;\\
    P_7=P_6\cup \{u(f(a))\leftarrow \naf u(f(a))\};\\
    P_8=P_7 \cup \{z(f(a))\leftarrow p(f(a))\};\\
    \vdots
  \end{array}
  \]
\end{exa}
}

Of course, we are particularly interested in those properties of
module sequences that are independent from the enumeration of GH.
We say that a ground subprogram $P'\subseteq \ground(P)$ is
\emph{downward closed}, if for each atom $A$ occurring in $\atom(P')$, the subprogram $P'$
contains all the rules $r\in \ground(P)$ such that $A\in head(r)$.

\begin{proposition}
\label{prop:propOfSplitSeq}
  Let $P$ be a disjunctive program. For all {module sequences}
  $P_1, P_2, ...$, for~$P$:
  \begin{enumerate}
    \item \label{prope1propOfSplitSeq} $\bigcup_{i\geq 1} P_i=\ground(P)$,
    \item for each $i\geq 1$ and $j\geq i$, $\atom(P_i)$ is a splitting 
      set of $P_j$, and $P_i=bot_{\atom(P_i)}(P_j)$,
    \item for each $i\geq 1$, $\atom(P_i)$ is a splitting 
      set of $P$, and $P_i=bot_{\atom(P_i)}(P)$,
    \item for each $i\geq 1$, $P_i$ is downward closed.
  \end{enumerate}   
\end{proposition}
This proposition follows easily from the definitions.  It shows that
each module sequence for $P$ consists of the bottom programs
corresponding to a particular splitting sequence $\langle
\atom(P_i)\rangle_{i<\omega}$ that depends on the underlying
enumeration of $GH$.  Roughly speaking, such sequences (whose length
is limited by $\omega$) constitute a \emph{normal form} for splitting
sequences and enjoy useful properties that are invariant with respect to the enumeration.  

\begin{definition}[Smoothness]
\label{smoothSeq}
  A transfinite sequence of sets $\langle X_\alpha\rangle_{\alpha<\mu}$
  is \emph{smooth} iff $X_0$ is finite and for each non-limit ordinal $\alpha + 1 < \mu$, 
  the difference $X_{\alpha + 1}\setminus X_\alpha$ is finite.
\end{definition}
Note that when $\mu=\omega$ (as in module sequences), smoothness implies 
that each $X_\alpha$ in the sequence is finite.  Finitely recursive programs 
are completely characterized by smooth module sequences:

\begin{theorem}
\label{newDefFinitelyRec}
For all disjunctive logic programs $P$, the following are equivalent:
\begin{enumerate}
\item\label{item1} $P$ is finitely recursive;
\item\label{item2} $P$ has a smooth module sequence (where each $P_i$ is finite);
\item\label{item3} all module sequences for $P$ are smooth.
\end{enumerate}
\end{theorem}
\begin{proof}
  \begin{description}
  \item [($\ref{item1} \Rightarrow \ref{item3}$)] Let $P$ be a finitely recursive program and 
  let $e= p_1,p_2,\ldots$ be any enumeration of $GH$.
  If $S=P_1,P_2,\ldots$ is the module sequence induced by the enumeration $e$
  then $S$ is smooth because, for each atom $p_i$ in $e$, the set 
  $\{ \, r\in \ground(P)\ \mid\ p_{i} \ depends\ on\ some\ A\in head(r) \, \}$   
  is finite, as $P$ is finitely recursive. Since this holds for 
  an arbitrary enumeration $e$, all module sequences for $P$ are smooth.  

  \item [($\ref{item3}\Rightarrow\ref{item2}$)] Trivial.

  \item [($\ref{item2}\Rightarrow\ref{item1}$)] Let $S=P_1,P_2,\ldots,$ be a smooth module sequence 
  for $P$ and let $p$ be an atom in $\ground(P)$. By Proposition~\ref{prop:propOfSplitSeq}.(\ref{prope1propOfSplitSeq}),
  there is a program $P_i$ in $S$ such that $p\in \atom(P_i)$. Moreover, $P_i$ is downward closed by definition of module sequence
  and it is finite because $S$ is smooth. Then $p$ depends only on finitely many ground atoms. Since $p$ has been arbitrarily chosen, the same holds for all ground atoms, therefore 
  $P$ is finitely recursive.
  \end{description}
\end{proof}

Smooth module sequences clearly correspond to smooth splitting
sequences of length $\omega$.  In particular, for each smooth module
sequence $\tup{P_i}_{i<\omega}$\,, $\tup{\atom(P_i)}_{i<\omega}$ is a
smooth splitting sequence.  Conversely, given a smooth splitting
sequence $\tup{U_i}_{i<\omega}$ and an arbitrary enumeration $p_1,
p_2, \ldots, p_i$, the resulting module sequence must necessarily be
smooth.  Suppose not; then some $p_i$ must depend on infinitely many
atoms.  Consequently, all the sets $U_j$ containing $p_i$ should be
infinite as well (a contradiction).  Note that in general a smooth
splitting sequence does not strictly correspond to a module sequence.
For example, the difference between two consecutive elements of a
splitting sequence may contain two atoms that do not depend on each
other, while this is impossible in module sequences by construction.

Using the above relationships between smooth module sequences and smooth splitting sequences of length $\omega$, the characterization of finitely recursive programs can be completed as follows, in terms of
standard splitting sequences:

\begin{corollary}
\label{smoothSplitSeq}
For all disjunctive programs  $P$, the following are equivalent:
\begin{enumerate}
\item $P$ is finitely recursive;
\item $P$ has a smooth splitting sequence of length $\mu\leq\omega$.
\end{enumerate}
\end{corollary}
\begin{proof}
  A straightforward consequence of Theorem~\ref{newDefFinitelyRec} and the correspondence between smooth module sequences and smooth splitting sequences of length $\mu\leq\omega$. 
\end{proof}

Note the asymmetry between Corollary~\ref{smoothSplitSeq} and
Theorem~\ref{newDefFinitelyRec}.  It can be explained by the generality of
splitting sequences: even if the underlying program is finitely
recursive, splitting sequences are not forced to be all smooth.  For
example, the finitely recursive program
\begin{eqnarray*}
  \mathit{even}(0)\\
  \mathit{even}(s(s(X))) & \leftarrow & \mathit{even}(X)\\
  \mathit{odd}(s(0))\\
  \mathit{odd}(s(s(X))) & \leftarrow & \mathit{odd}(X)\\
\end{eqnarray*}
has a non-smooth splitting sequence \tup{\{\mathit{even}(s^n(X)) \mid
  n \mbox{ even}\},\{\mathit{odd}(s^n(X)) \mid
  n \mbox{ odd}\}}\,.

Next we illustrate how module sequences provide an incremental characterization of the stable models of disjunctive logic programs.

Roughly speaking, the following theorem rephrases the splitting sequence theorem of \cite{lifschitz94splitting} 
in terms of module sequences.  The original splitting sequence theorem applies to sequences of disjoint program 
``slices'', while our theorem applies to monotonically increasing program sequences.  Since no direct proof of 
the splitting sequence theorem was ever published (only the proof of a more general result for default logic 
was published \cite{turner96splitting}), here we give a direct proof of our result.

\begin{theorem}[Module sequence theorem]
\label{ModelOfP_disj}
  Let $P$ be a disjunctive program and $P_1,$ $P_2, ...$ be a module sequence for $P$. 
  Then $M$ is a stable model of $P$ 
  iff there exists a sequence $M_1, M_2, ...$ such that :
  \begin{enumerate}
    \item \label{MiSMofPi_disj} for each $i\geq 1$, $M_i$ is a stable model of $P_i$,
    \item \label{MiInMi+1_disj} for each $i\geq 1$, $M_i=M_{i+1} \cap \atom(P_i)$,
    \item \label{MisUMi_disj} $M=\bigcup_{i\geq 1} M_i$.
  \end{enumerate}
\end{theorem}
\begin{proof}
  Let $M$ be a stable model of $P$.  
  Since $P_1, P_2, ...$ is a module sequence for $P$ then 
  for each $i\geq 1$, $\atom(P_i)$ is a splitting set of $P$ 
  and $P_i=bot_{\atom(P_i)}(P)$. Consider the sequence  of models $M_i=M\cap \atom(P_i)$\,, ($1\leq i < \omega$). By the splitting
  theorem \cite{lifschitz94splitting}, for each $i\geq 1$, $M_i$ is a stable model of $P_i$.
  Second, since $P_{i+1}\supseteq P_i$, we have $M_i = M\cap \atom(P_i) = (M\cap \atom(P_{i+1})) \cap \atom(P_i) = M_{i+1}\cap \atom(P_i)$\,.
  Finally, by Proposition~\ref{prop:propOfSplitSeq}.(\ref{prope1propOfSplitSeq}) we have $\bigcup_i M_i=M$.
  Then for each stable model $M$ of $P$ there exists a sequence 
  of finite sets of ground atoms that satisfies 
  properties \ref{MiSMofPi_disj}, \ref{MiInMi+1_disj} and \ref{MisUMi_disj}.     

  Conversely, let $P$ be a disjunctive logic program.  For the sake of readability, we assume without loss of generality that $P$ is ground. Suppose that there exists a sequence $M_1, M_2, ...$ that 
  satisfies properties \ref{MiSMofPi_disj}, \ref{MiInMi+1_disj} and \ref{MisUMi_disj}. 
  We have to prove that the set $M=\bigcup_{i\geq 1} M_i$ is a 
  stable model of $P$; equivalently,
  \begin{equation*}
    \bigcup_{i\geq 1} M_i\in\lm(P^M).
  \end{equation*}
  Properties \ref{MiInMi+1_disj} and \ref{MisUMi_disj} imply that for all $i\geq 1$, 
  $(M\cap \atom(P_i))=M_i$; consequently $P_i^M = P_i^{M_i}$ and by 
  Proposition~\ref{prop:propOfSplitSeq}.(\ref{prope1propOfSplitSeq}),
  \begin{equation}
  \label{eq:P^M=UP_i^M_i}
    P^M=\left(\bigcup_{i\geq 1} P_i\right)^M=\bigcup_{i\geq 1} P_i^{M}=\bigcup_{i\geq 1} P_i^{M_i}.
  \end{equation}

  First we prove that $M$ is a model of $P^M$, that is for each rule $r$ in $P^M$, if $body(r)\subseteq M$ then 
  $head(r)\cap M \neq \emptyset$. Let $r$ be any rule in $P^M$ such that $body(r)\subseteq M$. By equation~(\ref{eq:P^M=UP_i^M_i}), 
  there is an integer $i\geq 1$ such that $r\in P_i^{M_i}$.  Moreover, it is not hard to prove that properties \ref{MiInMi+1_disj}, \ref{MisUMi_disj} and $body(r)\subseteq M$ imply $body(r)\subseteq M_i$.  Now, since $M_i$ is a stable model 
  of $P_i$ and $body(r)\subseteq M_i$, we have $head(r)\cap M_i \neq \emptyset$. It follows immediately that $head(r)\cap M \neq \emptyset$.  Since this holds for any $r\in P^M$, we conclude that $M$ is a model of $P^M$.

  We are left to show that $M$ is a minimal model for $P^M$. Suppose that $P^M$ has a model $M'\subset M$. Let $p\in (M\setminus M')$ and let $i$ be an integer 
  such that $p\in \atom(P_i)$. Since $P_i^M=P_i^{M_i}$ is a bottom program for $P^M$ then $M'\cap\atom(P_i)$ is a 
  model for $P_i^{M_i}$ and it is strictly contained in $M_i$, but this is a contradiction because by hypothesis
  $M_i$ is a minimal model of $P_i^{M_i}$.   
\end{proof}

The module sequence theorem (respectively, the splitting sequence
theorem) suggests a relationship between the consistency 
of a program $P$ and the consistency of each step in $P$'s module
sequences (respectively, the sequence of program slices induced by
$P$'s splitting sequences). To clarify this point we introduce another 
invariant property of module sequences.

\begin{definition}
  A module sequence $S=P_1, P_2, ...$ for a disjunctive program $P$ 
  is \emph{inconsistent} if for some $i<\omega$,
  $P_i$ is inconsistent; otherwise $S$ is \emph{consistent}.  
\end{definition}

\begin{proposition}
\label{Pinconsistent}
  If a disjunctive program $P$ has an inconsistent
  module sequence then $P$ is inconsistent.
\end{proposition}
\begin{proof}
  Suppose that $P$ has an inconsistent module sequence
  $P_1, P_2, ...$, that is, some $P_i$ in the sequence is inconsistent. It follows that $P$ has an inconsistent bottom program and hence $P$ is inconsistent
  by the splitting theorem. 
\end{proof}


\begin{theorem}
\label{inconsPiffSincons}
  Let $S=P_1, P_2, ...$ be a module sequence for a 
  disjunctive program $P$. If $S$ is inconsistent 
  then each module sequence for $P$ is inconsistent.
\end{theorem} 
\begin{proof}
  Let $S=P_1, P_2, ...$ be an inconsistent module sequence for $P$
  induced by the enumeration $p_1, p_2, ...$ of $GH$ and let
  $i$ be the least index such that $P_i$ is inconsistent.
  Let $S'=P'_1, P'_2, ...$ be any module sequence for $P$ 
  induced by the enumeration $p'_1, p'_2, ...$ of $GH$.
  Since $i$ is finite, there exists a finite $k$ such that 
  $\{p_1, p_2, ..., p_i\}\subseteq \{p'_1, p'_2, ..., p'_k\}$.
  So, by construction, $P_i\subseteq P'_k$ and then $\atom(P_i)\subseteq \atom(P'_k)$.
  Moreover, by definition, $P_i$ is downward closed, therefore $P_i=bot_{\atom(P_i)}(P'_k)$.
  Since $P_i$ is inconsistent, $P'_k$ is inconsistent (by the splitting theorem) 
  and hence $S'$ is inconsistent, too. 
\end{proof}
In other words, for a given program $P$, either all module sequences 
are inconsistent, or they are all consistent.  In particular,  if $P$ is consistent, 
then every member $P_i$ of any module sequence for $P$ must be consistent.  
The converse property would allow to define a procedure for
enumerating the stable models of $P$ (as shown in the following
sections).  Unfortunately, even if each step in a module sequence is
consistent, the entire program $P$ is not necessarily consistent, as shown by the following example.

\begin{example}
\label{counterexa}
  As a preliminary step, consider the following program $P_f$ (due to
  Fages \cite{Fag94}):
 
  \begin{align*}
    q(X) &\leftarrow q(f(X)).\\
    q(X) &\leftarrow \naf q(f(X)).\\
    r(0).
  \end{align*}
  
  The third rule is only needed to introduce the constant $0$ into the
  program's language.

  This program is inconsistent. To see this, note that -- roughly speaking -- the first two
  rules in $P_f$ are classically equivalent to
  \begin{equation*}
    q(X)\leftarrow q(f(X)) \lor \naf q(f(X)) \,.
  \end{equation*}
  Since the body is a tautology and the stable models of a program are
  also classical models of the program (if $\naf$ is interpreted as $\neg$), we have that every stable model of
  $P_f$ should satisfy all ground instances of $q(X)$.  However, the
  Gelfond-Lifschitz transformation with respect to\ such a model would contain
  only the first and the third program rules, and hence the least
  model of the transformation would contain no instance of $q(X)$.  It
  follows that $P_f$ is inconsistent (it has no stable models).
Now consider the following extension $P$ of $P_f$:
  \begin{enumerate}
    \item \label{r1} $q(X)\leftarrow q(f(X)), p(X).$
    \item \label{r2} $q(X)\leftarrow \naf q(f(X)), p(X).$
    \item \label{r3} $r(0).$

    \item \label{r4} $p(X)\leftarrow \naf p'(X).$
    \item \label{r5} $p'(X)\leftarrow \naf p(X).$

    \item \label{r6} $c(X)\leftarrow \naf c(X), \naf p(X).$
  \end{enumerate}
  The program $P$ is inconsistent, too. To verify it, suppose that $M$ is a stable model of
  $P$. By rules \ref{r4} and \ref{r5}, for all ground instances of $X$,
  exactly one of $p(X)$ and $p'(X)$ is true in $M$. However, if $p(X)$
  is false, then rule \ref{r6} produces an inconsistency due to the
  odd-cycle involving $c(X)$. It follows that all ground
  instances of $p(X)$ must be true in $M$. But then
  rules \ref{r1}, \ref{r2} and \ref{r3} become equivalent to program
  $P_f$ and prevent $M$ from being a stable model, as explained above.
  So $P$ is inconsistent.
  
  Next, consider the enumeration 
  $e= r(0),$ $q(0),$ $p(0),$ $p'(0),$ $c(0),$ $q(f(0)),$ 
  $p(f(0)),$ $p'(f(0)),$ $c(f(0)),$ $...$\,, 
  of the set $GH$.  This enumeration induces the following module
  sequence for $P$ (where the expression $[X/t]$ denotes the
  substitution mapping $X$ onto~$t$):

  \begin{center}
    \renewcommand{\arraystretch}{1.25}
  \begin{tabular}{lll}
    $P_0$ & $=\{r(0)\}$ & \\
    $P_1$ & $=P_0 \cup \bigcup_{k < \omega}\{$ & $q(X)\leftarrow q(f(X)), p(X),$\\ 
      & & $q(X)\leftarrow \naf q(f(X)), p(X),$ \\ 
      & & $p(X)\leftarrow \naf p'(X),$\\
      & & $p'(X)\leftarrow \naf p(X) \, \} \, [X/f^k(0)]$\\
    $P_{i+1}$ & \multicolumn{2}{l}{$=P_i \cup \{c(X)\leftarrow \naf c(X), \naf p(X)\} \, [X/f^{i-1}(0)] \qquad (i\geq 1)\,.$}
  \end{tabular}
  \end{center}
  
  Note that $M_0=\{r(0)\}$ is a stable model of $P_0$
  and for each $i\geq 1$ and $k\geq i-2$  
  
  \begin{align*}
    M_i^k=\{ & r(0), p(f^0(0)),p(f^1(0)),p(f^2(0)), \ldots,p(f^{k}(0)),\\
    & p'(f^{k+1}(0)),p'(f^{k+2}(0)),\ldots,p'(f^{k+j}(0)), \ldots \\
    & q(f^0(0)),q(f^1(0)),q(f^2(0)),\ldots,q(f^{k}(0))\} 
  \end{align*} 
  
  is a stable model of $P_i$.  Therefore, each $P_i$ is consistent 
  while $\bigcup_i P_i=\ground(P)$ is inconsistent. 
  This happens because for each stable model $M$ of $P_1$ there exists
  a $P_j$ $(j>1)$ such that $M$ is not the bottom part of any stable
  model of $P_j$.  Intuitively,  $M$ has been ``eliminated'' at step
  $j$.  In this example $P_1$ has infinitely many stable models, and
  it turns out that no finite step eliminates them all.  
  Consequently, each $P_i$ in the module sequence is consistent, but the entire program is not.
\end{example}

Note that $P$ is not finitely recursive because, for each grounding
substitution $\sigma$, $q(X)\sigma$ depends on the infinite set of
ground atoms $\{\, q(f(X))\sigma,\ q(f(f(X)))\sigma,\ ... \, \}$ (due
to rules (\ref{r1}) and (\ref{r2})).
In the following section we are going to prove that finitely recursive
programs are not affected by the problem illustrated in
Example~\ref{counterexa}, that is, they enjoy the converse of
Theorem~\ref{inconsPiffSincons}.  This property will be used to design an
effective enumeration procedure for their stable models.

	\section{Compactness property of disjunctive finitely
	  recursive programs} 
	\label{sec:compactness}

Here we prove that the compactness theorem proved in
\cite{DBLP:conf/iclp/BaseliceBC07} for normal finitely recursive
programs actually holds for all disjunctive finitely recursive
programs.

The first step is to prove the converse of
Theorem~\ref{inconsPiffSincons} for all finitely recursive programs.

\begin{theorem}
\label{ConsSplitSeqPCons}
  For all disjunctive finitely recursive programs $P$, if some module
  sequence for $P$ is consistent, then $P$ is consistent.
\end{theorem}
\begin{proof}
  Let $S$ be any module sequence for $P$.  If $S$ is consistent,
  then each module $P_i$ in $S$ has a nonempty set of stable
  models. It suffices to prove that there exists a sequence $M_1, M_2,
  ...$ of stable models of $P_1, P_2, ...$, respectively, that
  satisfy the properties of Theorem~\ref{ModelOfP_disj}, because
  this implies that $M=\bigcup_i M_i$ is a stable model of $P$.

  We call a stable model $M_i$ of $P_i$ \emph{``bad''} if there 
  exists a $k>i$ such that no stable model $M_k$ of $P_k$ extends $M_i$, 
  \emph{``good''} otherwise. We say that $M_k$ extends $M_i$ if
  $M_k\cap \atom(P_i)=M_i$.

  \emph{Claim 1}: Each 
  $P_i$ must have at least one \emph{``good''} stable model.

  To prove the claim, suppose that there exists an $i$ such that all stable
  models of $P_i$ are \emph{``bad''}.  Since $P_i$ is a finite program,
  it has a finite number $M_{i,1},...,M_{i,r}$ of stable models.  By
  assumption, for each $M_{i,j}$ there is a program $P_{k_j}$ none
  of whose stable models extends $M_{i,j}$.  Let
  $k=max\{k_1,...,k_r\}$; clearly, no stable model of $P_k$ extends any stable
  model of $P_i$, and this is a contradiction because $P_k$, by
  hypotheses, has at least one stable model $M_k$ and by the splitting
  theorem, $M_k$ must extend a stable model of $P_i$.  This proves
  Claim 1.

  \emph{Claim 2}: Each \emph{``good''} stable model $M_i$ of $P_i$ is extended by some 
  \emph{``good''} stable model $M_{i+1}$ of $P_{i+1}$.

  Suppose not. Then, none of the stable models $M_{i+1,1},\ldots,M_{i+1,r}$ of
  $P_{i+1}$ that extend the good stable model $M_i$ of $P_i$ is good.  This implies (by analogy with the
  proof of Claim 1) that there exists a module $P_k$ ($k>i+1$) none of whose
  stable models extends any of $M_{i+1,1},\ldots,M_{i+1,r}$.  It follows that none of $P_k$'s stable
  models can extend $M_i$, and this contradicts the hypothesis that $M_i$
  is good.  

  From the two claims it follows immediately that there exists an infinite sequence $M_1,$ $M_2, ...$ that
  satisfies properties \ref{MiSMofPi_disj} and \ref{MiInMi+1_disj} of
  Theorem~\ref{ModelOfP_disj} and hence the union $M=\bigcup_i M_i$ is
  a stable model of $P$.  
\end{proof}
Note that in Example~\ref{counterexa}, module $P_1$ is infinite and has
infinitely many stable models, all of which are ``bad''.  Each of them
is eliminated at some step, but no finite step eliminates them all,
which is why that module sequence is consistent although the entire
program $P$ is not.

Theorem~\ref{ConsSplitSeqPCons} can be extended to all smooth splitting sequences
with length~$\omega$:

\begin{corollary}
  \label{thm:split-seq-version}
  Let $\langle U_\alpha\rangle_{\alpha < \omega}$ be a 
  smooth splitting sequence for a disjunctive program $P$.
  Then $P$ is consistent iff for all $\alpha < \omega$, $bot_{U_\alpha}(P)$
  is consistent.
\end{corollary}

\begin{proof}
A straightforward consequence of the correspondence between module and splitting sequences.  
\end{proof}

The restriction to sequences with length $\omega$ is essential to
derive the above corollary, which is not valid otherwise, as shown by
the following example.

\begin{example}
Let $P$ be the following program, where rule~\ref{r20} has the role of
creating an infinite Herbrand domain:
  \begin{enumerate}
    \item \label{r20} $r(f(0)).$
    \item \label{r21} $p(X)\leftarrow \naf q(X).$
    \item \label{r22} $q(X)\leftarrow \naf p(X).$

    \item \label{r24} $some\_q\leftarrow q(X).$
    \item \label{r25} $f \leftarrow \naf f,\ \naf some\_q.$

    \item \label{r26} $c(X)\leftarrow \naf c(X), q(X).$
  \end{enumerate}
This program is inconsistent for the following reasons: For all ground
instances of $X$, rules~\ref{r21} and \ref{r22} force exactly one
of $p(X)$ and $q(X)$ to be true.  If no instance of $q(X)$ is true,
then rules~\ref{r24} and \ref{r25} create a contradiction
by ``activating'' the odd-cycle involving $f$.  However, if some
instance  of $q(X)$ is true, then  rule~\ref{r26} generates a contradiction
by ``activating'' the odd-cycle involving $c(X)$.  It follows that $P$
has no stable models.

However, $P$ has a smooth splitting sequence with length $2\omega$
whose bottom programs are all consistent:
\begin{align*}
  U_0 &= \{ \, p(0),\ q(0),\ r(0) \, \} \\
  U_i &= U_{i-1} \cup \{ \, p(f^i(0)),\ q(f^i(0)),\ r(f^i(0))\, \} && (0<i<\omega)\\
  U_\omega &= \bigcup_{i<\omega} U_i \\
  U_{\omega+1} &= U_\omega \cup \{ \, some\_q,\ f,\ c(0) \, \} \\
  U_{\omega+j+1} &= U_{\omega+j} \cup \{ \, c(f^j(0)) \, \}
  && (0\leq j<\omega).
\end{align*}
In particular, 
\begin{enumroman}
\item   $bot_{U_\omega}(P)$ has infinitely many stable models,
  one for each choice between $p(X)$ and $q(X)$, for all instances of
  $X$;
\item   $bot_{U_{\omega+1}}(P)$ keeps all the stable models where at least
  one instance of $q(X)$ is true;
\item  $bot_{U_{\omega+j+1}}(P)$ keeps only those stable models where the
  first true instance of $q(X)$ is $q(f^k(0))$ with $k>j$.
\end{enumroman}
\end{example}

Now we are ready to extend the compactness property of finitary normal
programs to all disjunctive finitely recursive programs.

\begin{definition}
\label{def:unstableKernel}
  An \emph{unstable kernel} for a disjunctive program $P$ is a set $K\subseteq \ground(P)$
  with the following properties:
  \begin{enumerate}
  \item \label{cond1Kunstable} $K$ is downward closed.
  \item \label{cond2Kunstable} $K$ has no stable model. 
  \end{enumerate}
\end{definition}

\begin{theorem}[Compactness]
\label{Pcompactness}
  A disjunctive finitely recursive program $P$ has no stable model 
  iff it has a finite unstable kernel. 
\end{theorem}
\begin{proof}  
  By Proposition~\ref{Pinconsistent} and Theorem~\ref{ConsSplitSeqPCons}, $P$ has no
  stable model iff it has
  an inconsistent module sequence.
  So, let $P_1, P_2,..., P_n, ...$ be an inconsistent module sequence for $P$
  and choose an index $i\geq 1$ such that $P_i$ is inconsistent. By 
  Proposition~\ref{prop:propOfSplitSeq}, 
  $P_i\subseteq \ground(P)$; moreover, 
  $P_i$ is downward closed. Then $P_i$ is an unstable
  kernel for $P$. Moreover, by Theorem~\ref{newDefFinitelyRec},
  $P_i$ is finite. 
\end{proof}

	\section{Reasoning with disjunctive finitely recursive
  programs}
\label{sec:complexity}

By taking an effective enumeration of the set $GH$ of ground head atoms,
one can effectively compute each element of the corresponding module sequence. 
Let us call \textsc{construct}\,$(P,i)$ an effective procedure that, 
given a finitely recursive program $P$ and an index $i$, 
returns the ground program $P_i$, and let $\SM(P_i)$ 
be an algorithm that computes the finite set of the finite stable models of $P_i$:

\begin{theorem}
\label{Psemidec}
  Let $P$ be a disjunctive finitely recursive program. Deciding whether $P$ is inconsistent 
  is semidecidable. 
\end{theorem}
\begin{proof}
  Given a module sequence $P_1, P_2, ..., P_n, ...$ for the program $P$,
  consider the algorithm \textsc{consistent}\,$(P)$.

  \begin{algorithm}
    {\bf Algorithm} \textsc{consistent}\,$(P)$
    \label{alg1}
    \begin{algorithmic}[1]
      \STATE $i=0$;
      \STATE $answer=$ $\true$;
      \REPEAT 
        \STATE $i=i+1$;
        \STATE $P_i=$\textsc{construct}\,$(P,i)$;
        \IF {$\SM(P_i)= \emptyset$}
          \STATE $answer =$ $\false$;
        \ENDIF
      \UNTIL {$\neg answer$ OR $P_i=\ground(P)$}
      \RETURN $answer$;
    \end{algorithmic}
  \end{algorithm}
 
  By Proposition~\ref{Pinconsistent} and Theorem~\ref{ConsSplitSeqPCons}, 
  $P$ is inconsistent iff
  there exists an $i\geq 1$ such that $P_i$ is inconsistent
  (note that we can always check the consistency of $P_i$ because 
  $P_i$ is finite). Then, the algorithm returns $\false$ iff
  $P$ is inconsistent.

  Note that if $\ground(P)$ is infinite then any module sequence
  for $P$ is infinite and the algorithm \textsc{consistent}\,$(P)$ 
  terminates iff $P$ is not consistent.  
\end{proof}

Next we deal with skeptical inference.  
Recall that a closed first order formula $F$ is a skeptical consequence of $P$ iff 
$F$ is satisfied (according to classical semantics) by all the stable models of $P$.

\begin{theorem}
\label{finiteSkepInf}
  Let $P$ be a disjunctive finitely recursive program and $P_1,$ $P_2,
  ...$ be a module sequence for $P$.  A ground formula $F$ in the
  language of $P$ is a skeptical consequence of $P$ iff there exists a
  finite $k\geq 1$ such that $F$ is a skeptical consequence of $P_k$ and
  $\atom(F)\subseteq \atom(P_k)$.%
\end{theorem}
\begin{proof}
  Let $h$ be the least integer such that $\atom(F)\subseteq \atom(P_h)$
  (note that there always exists such an $h$ because $\atom(F)$
  is finite).
  Suppose that there exists a $k\geq h$ such that\ $F$ is a skeptical consequence of $P_k$. 
  Since $P_k$ is a bottom program for $P$, then each stable model $M$ of $P$ 
  extends a stable model $M_k$ of $P_k$ and then
  satisfies $F$ (here the assumption that $\atom(F)\subseteq \atom(P_k)$ is essential to conclude that $M$ and $M_k$ agree on the truth of $F$). So, $F$ is a skeptical consequence of $P$.  This proves the
  ``if'' part.

  Now suppose that, for each $k\geq h$, $F$ is not 
  a skeptical consequence of $P_k$. 
  This implies that each $P_k$ 
  is consistent (hence $P$ is consistent) and, moreover, 
  the set $S$ of all the stable models of $P_k$ that falsify
  $F$ is not empty.  

  Note that $S$ is finite because
  $P_k$ is finite (as $P$ is finitely recursive).
  So, if all the models in $S$ are ``\emph{bad}'' (cf.\ the proof of
  Theorem~\ref{ConsSplitSeqPCons}), 
  then there exists a finite integer $j>k$ such that\ no model 
  of $P_j$ contains any model of $S$.  Consequently, $F$ is a skeptical
  consequence of $P_j$---a contradiction. 

  Therefore at least one of these model must be \emph{good}.  Then there must
  be a model $M$ of $P$ that contains this ``\emph{good}'' model
  of $P_k$, and hence $F$ is not a skeptical consequence of $P$.       
\end{proof}
The next theorem follows easily.

\begin{theorem}
\label{th20}
  Let $P$ be a disjunctive finitely recursive program.
  For all ground formulas $F$,
  the problem of deciding whether $F$ is 
  a skeptical consequence of $P$ 
  is semidecidable.
\end{theorem}
\begin{proof}
  Given a module sequence $P_1, P_2, ..., P_n, ...$ for the program $P$,
  consider the algorithm \textsc{skeptical}\,$(P,F)$.

  \begin{algorithm}
    {\bf Algorithm} \textsc{skeptical}\,$(P,F)$
    \label{algSkep}
    \begin{algorithmic}[1]
      \STATE $answer=$ $\false$;
      \STATE $i=0$;
      \REPEAT 
        \STATE $i=i+1$;
        \STATE $P_i=$\textsc{construct}\,$(P,i)$;
      \UNTIL {$\atom(F)\subseteq \atom(P_i)$}
      \REPEAT 
        \IF {$\SM(P_i)= \emptyset$ OR $P_i$ skeptically entails $F$}
          \STATE $answer =$ $\true$;
        \ELSE
          \STATE $i=i+1$;
          \STATE $P_i=$\textsc{construct}\,$(P,i)$;          
        \ENDIF
      \UNTIL {$answer$ OR $P_i=P$}
      \RETURN $answer$;
    \end{algorithmic}
  \end{algorithm}
  
  For each $P_i$ such that $\atom(F)\subseteq \atom(P_i)$, 
  the algorithm \textsc{skeptical}\,$(P,$ $F)$ checks if $F$ is a 
  skeptical consequence of $P_i$.   
  Since $P_i$ is finite, we can always decide if $F$ is a 
  skeptical consequence of $P_i$.
  So, by Theorem~\ref{finiteSkepInf}, the algorithm returns $\true$ 
  iff $F$ is a skeptical consequence of $P$.

  Note that if $\ground(P)$ is infinite then any module sequence for $P$ 
  is infinite and the algorithm \textsc{skeptical}\,$(P,F)$ teminates iff 
  $F$ is a skeptical consequence of $P$.
 \end{proof}

For a complete characterization of the complexity of ground queries and inconsistency checking, 
we are only left to prove that the above upper bounds are tight.
Actually, we prove slightly stronger lower bounds, that hold even for \emph{normal} finitely recursive
programs.

\begin{theorem}
\label{InconsPreComplete}
  Deciding whether a normal finitely recursive program $P$ is
  inconsistent is r.e.-hard.    
\end{theorem}
\begin{proof}  
  The proof is by reduction of the
  problem of skeptical inference of a quantified formula over a 
  finitary normal program (proved to be r.e.-complete in
  \cite[Corollary~23]{DBLP:journals/ai/Bonatti04}) to the problem of inconsistency checking 
  over a normal finitely recursive program.
  
  Let $P$ be a finitary program and $\exists F$ be a closed existentially quantified formula.
  Let $((l_{11} \vee$ $l_{12} \vee ...) \wedge$ $(l_{21} \vee$ $l_{22} \vee ...)\wedge ...)$
  be the conjunctive normal form of $\neg F$. Then
  $\exists F$ is a skeptical consequence of $P$ iff the program $P\cup C$ 
  is inconsistent, where
    \[
       C=
        \left \{
         \begin{array}{l@{\quad}p{9em}}
           p_1(\vec{x_1})\leftarrow \naf l_{11}, \naf l_{12},..., \naf p_1(\vec{x_1})  
          \\
           p_2(\vec{x_2})\leftarrow \naf l_{21}, \naf l_{22},..., \naf p_2(\vec{x_2})             
          \\
           \quad \qquad \vdots
         \end{array}
        \right \},
    \]
  $p_1$, $p_2$, ... are new atom symbols not occurring in $P$ or $F$, and $\vec{x_i}$ 
  is the vector of all variables occurring in $(l_{i1} \vee$ $l_{i2} \vee ...)$.  
  Note that $P\cup C$ is a normal finitely recursive program.
  
  The constraints in $C$ add no model to $P$, but they
  only discard those models of $P$ that satisfy $F\theta$ (for some substitution $\theta$). 
  So, let $\SM(P)$ be the set of stable models of $P$. Then each model in 
  $\SM(P\cup C)$ 
  satisfies $\forall \neg F$. 
  $\SM(P\cup C)=\emptyset$ (that is $P\cup C$ is inconsistent) iff either $\SM(P)=\emptyset$ 
  or all stable models of $P$ satisfy $\exists F$.
  Then $\SM(P\cup C)=\emptyset$ iff $\exists F$ is a skeptical consequence of $P$.    
\end{proof}

\begin{theorem}
\label{th:skep-r-e-complete}
  Deciding whether a normal finitely recursive program $P$ skeptically 
  entails a ground formula $F$ is r.e.-hard.  
\end{theorem}
\begin{proof}
  The proof is by
  reduction of inconsistency checking 
  for normal finitely recursive programs
  to the problem of skeptical inference of a ground formula from a 
  normal finitely recursive program.
   
  Let $P$ be a normal finitely recursive program and $q$ be a new 
  ground atom that doesn't occur in $P$. Then, $P$ is 
  inconsistent iff $q$ is a skeptical consequence of $P$.
  Since $q$ occurs in the head of no rule of $P$,
  $q$ cannot occur in a model of $P$. So, $P$ skeptically 
  entails $q$ iff $P$ has no stable model. 
\end{proof}

\begin{corollary}
\label{credInf-co-r-e}
  Deciding whether a disjunctive finitely recursive program $P$ 
  credulously entails a ground formula $F$ is co-r.e. complete.
\end{corollary}
\begin{proof}
  The proof follows immediately from Theorems~\ref{th20} and~\ref{th:skep-r-e-complete} 
  and from the fact that a ground formula
  $F$ is a credulous consequence of $P$ iff $\neg F$ is not
  a skeptical consequence of $P$.
\end{proof}

\section{Skeptical resolution for finitely recursive normal programs}
\label{sk-resolution}

In this section we extend the work in
\cite{DBLP:journals/jar/Bonatti01,DBLP:journals/ai/Bonatti04} by
proving that skeptical resolution (a top-down calculus which is known
to be complete for Datalog and normal finitary programs under the
skeptical stable model semantics) is complete also for the class of
finitely recursive normal programs.  Skeptical resolution has several
interesting properties.  For example, it does not require the input
program $P$ to be instantiated before reasoning (unlike the major
state-of-the-art stable model reasoners), and it can produce
nonground (i.e., universally quantified) answer substitutions.  The
goal-directed nature of skeptical resolution makes it more interesting
than the naive algorithms illustrated in Section~\ref{sec:complexity}.

We are not describing all the formal details of the calculus here---the reader is referred to \cite{DBLP:journals/jar/Bonatti01}.  
Skeptical resolution is based on 
\emph{goals with hypotheses} (\emph{h-goals} for short) which are pairs $(G \mid H)$ 
where $H$ and $G$ are finite sequences of literals.
Roughly speaking, the answer to a query 
$(G \mid H)$ should be \emph{yes} if $G$ holds in all the stable 
models that satisfy $H$. Hence $(G \mid H)$ has the same meaning 
in answer set semantics as the implication $(\bigwedge G\leftarrow \bigwedge H)$.
Finally, a \emph{skeptical goal}
(\emph{s-goal} for short) is a finite sequence of h-goals.

The calculus consists of five inference rules:

\begin{description}
\item[Resolution.] This rule may take two forms; a literal can be
unified with either a program rule or a hypothesis.  First suppose that
$L_i$ is an atom, $A\leftarrow B_1,\ldots,B_k$ is a standardized apart
variant of a rule of $P$, and $\theta$ is the \emph{mgu} of $L_i$ and
$A$.  Then the following is an instance of the rule.
 \[
\frac{	\Gamma\
	(L_1\ldots L_{i-1},\, L_i,\, L_{i+1}\ldots L_n \mid H)\ 
	\Delta}%
{	\left[\Gamma\ 
	(L_1\ldots L_{i-1},\, B_1,\ldots,B_k,\, L_{i+1}\ldots L_n
	 \mid H)\ 
	\Delta \right] \theta} \,.
\]
Next, let $L_i$ be a (possibly negative) literal, let $L'$ be a
hypothesis, and let $\theta$ be the \emph{mgu} of $L_i$ and $L'$. 
Then the following is an instance of the rule.
 \[
\frac{	\Gamma\
	(L_1\ldots L_{i-1},\, L_i,\, L_{i+1}\ldots L_n \mid H,L')\ 
	\Delta}%
{	\left[\Gamma\ 
	(L_1\ldots L_{i-1},\, L_{i+1}\ldots L_n \mid H,L')\ 
	\Delta \right] \theta} \,.
\]

\item[Contradiction.] This rule tries to prove $(G\mid H)$ ``vacuously'', by
showing that the hypotheses $H$ cannot be satisfied by any stable model of $P$.  Hereafter $\bar L=\naf A$ if $L$ is an atom $A$, and $\bar L=A$ if $L=\naf A$.
\[
\frac{	\Gamma\ (G\mid H,L)\ \Delta }%
{	\Gamma\ (\bar L \mid H,L)\ \Delta } \,.
\]

\item[Split.]
Essentially, this rule is needed to compute floating conclusions and
discover contradictions.  It splits the search space by introducing two
new, complementary hypotheses.  Let $G_0$ be the \emph{restart goal} (i.e.\ the left-hand side of the first h-goal of the derivation), $L$ be an arbitrary
literal and $\sigma$ be the composition of the \emph{mgu}s previously
computed during the derivation; the Split rule is:
 \[
\frac{	\Gamma\ (G\mid H)\ \Delta }%
{	\Gamma\ (G\mid H,L)\ (G_0\sigma\mid H,\bar L)\ \Delta } \,,
\]

\item[Success.]  This is a structural rule that removes h-goals once they have been successfully proved.  As usual, $\Box$ denotes the empty goal.
\[
\frac{	\Gamma\ (\Box\mid H)\ \Delta }%
{	\Gamma\ \Delta } \,.
\]
\end{description}

We are left to illustrate the last rule of the calculus, that models negation as failure.  In order to abstract away the details of the computation of failed facts, the rule is expressed in terms of so-called \emph{counter-supports}, that in turn are derived from the standard notion of \emph{support}.  Recall that a support for a ground atom $A$ is a set of negative literals obtained by applying SLD resolution to $A$ with respect to\ the given program $P$ until no positive literal is left in the current goal (the final, negative goal of the SLD derivation is a support for $A$).

\begin{definition}[\cite{DBLP:journals/jar/Bonatti01}]
  Let $A$ be a ground atom. A \emph{ground counter-support}
  for $A$ in a program $P$ is a set of atoms $K$ with the 
  following properties:
  \begin{enumerate}
    \item For each support $S$ for $A$, there 
    exists $\naf B\in S$ such that $B\in K$.
    \item For each $B\in K$, there exists a support 
    $S$ for $A$ such that $\naf B\in S$.
  \end{enumerate}  
\end{definition}

In other words, the first property says that $K$ contradicts 
all possible ways of proving $A$, while the second property 
is a sort of relevance property. 
Informally speaking, the failure rule of skeptical resolution says that if all atoms in a counter-support are true, then all
attempts to prove $A$ fail, and hence $\naf A$ can be concluded.

Of course, in general, counter-supports are not
computable and may be infinite (while skeptical derivations and their
goals should be finite). 

In \cite{DBLP:journals/jar/Bonatti01}
the notion of counter-support is generalized to non ground 
atoms in the following way:
\begin{definition}
  A (generalized) \emph{counter-support} for a ground atom $A$ is a pair
  $\langle K,\theta\rangle$ where $K$ is a set of atoms 
  and $\theta$ a substitution, such that for all grounding 
  substitutions $\sigma$, $K\sigma$ is a ground 
  counter-support for $A\theta\sigma$.
\end{definition}

The actual mechanism for computing counter-supports can be abstracted by means of a suitable function $\CounterSupp$, mapping each (possibly nonground) 
atom $A$ onto a set of \emph{finite} generalized counter-supports for $A$.
The underlying intuition is that function $\CounterSupp$ captures 
all the negative inferences that can actually be computed 
by the chosen implementation. 
Now negation-as-failure can be axiomatized as follows:

\begin{description}
\item[Failure.] Suppose that $L_i=\naf A$\,, and
$\tup{\{B_1,\ldots,B_k\},\theta}\in \CounterSupp(A)$.  Then the following is
an instance of the Failure rule.
\[
\frac{	\Gamma\
	(L_1\ldots L_{i-1},\, L_i,\, L_{i+1}\ldots L_n \mid H)\ 
	\Delta}%
{	\left[\Gamma\ 
	(L_1\ldots L_{i-1},\, B_1,\ldots,B_k,\, L_{i+1}\ldots L_n
	 \mid H)\ 
	\Delta \right] \theta} \,.
\]
\end{description}

To achieve completeness for 
the nonground skeptical resolution calculus, we need the 
negation-as-failure mechanism to be complete in the following sense.

\begin{definition}
  The function $\CounterSupp$ is \emph{complete}
  iff for each atom $A$, for all of its ground instances
  $A\gamma$, and for all ground counter-supports $K$ for 
  $A\gamma$, there exist $\langle K', \theta\rangle\in \CounterSupp(A)$
  and a substitution $\sigma$ such that $A\theta\sigma=A\gamma$ and
  $K'\sigma=K$. 
\end{definition} 

A \emph{skeptical derivation from P and $\CounterSupp$ with restart goal $G_0$} 
is a (possibly infinite) sequence of s-goals $\Gamma_0$,$\Gamma_1$,..., where
each $\Gamma_{i+1}$ is obtained from $\Gamma_i$ through one of the five 
rewrite rules of the calculus.   A skeptical derivation is \emph{successful} if its last s-goal is
empty; in this case we say that the first s-goal has a
successful skeptical derivation from $P$. 

\begin{example}
Let $P$ be
\begin{enumerate}
\item \label{r2.1}	$p(X) \leftarrow  \naf q(X)$ \\
\item \label{r2.2}	$q(X) \leftarrow  \naf p(X)$ \\
\item \label{r2.3}	$r(f(X)) \leftarrow  \naf p(X)$ \\
\item \label{r2.4}	$r(f(X)) \leftarrow  \naf q(X)$ 
\end{enumerate}
For all ground terms $t$, the literal $\naf p(t)$ is the unique support of $q(t)$.  
Therefore, we can set $\CounterSupp(q(X)) = \{ \tup{p(X),\varepsilon} \}$ (where $\varepsilon$ denotes the empty substitution), 
since the truth of $p(X)$ suffices to block all derivations of $q(X)$, for all possible values of $X$ 
(the issue of how to compute \CounterSupp will be briefly discussed at the end of this section).  
The following is a successful derivation of $(r(Y) \mid \emptyset)$ from $P$ with answer substitution $[Y/f(X)]$, 
showing that for all $X$, 
$r(f(X))$ is a skeptical consequence of $P$.
\[
\begin{array}{r@{\quad}p{16.5em}}
 (r(Y) \mid \emptyset) &      \\
 (\naf p(X) \mid \emptyset) & by resolution with \ref{r2.3}; it binds $Y$ to $f(X)$;     \\
 (\naf p(X) \mid \naf p(X)) \, (r(f(X)) \mid p(X)) & by the splitting rule;     \\
 (\Box \mid \naf p(X)) \, (r(f(X)) \mid p(X)) & by resolution with the hypothesis;     \\
 (r(f(X)) \mid p(X)) & by the success rule;     \\
 (\naf q(X) \mid p(X)) & by resolution with \ref{r2.4};     \\
 (p(X) \mid p(X)) & by the failure rule using \tup{p(X),\varepsilon};     \\
 (\Box \mid p(X)) & by resolution with the hypothesis;     \\
 \Box & by the success rule.
\end{array}
\]
\end{example}

Skeptical resolution is sound for \emph{all} normal programs and counter-support calculation mechanisms, as stated in the following theorem.

\begin{theorem}[Soundness, \cite{DBLP:journals/jar/Bonatti01}]
  Suppose that an \emph{s-goal} $(G \mid H)$ has a successful skeptical derivation 
  from a normal program $P$ and $\CounterSupp$ with restart goal $G$ and answer substitution 
  $\theta$. Then, for all grounding substitution $\sigma$, all 
  the stable models of $P$ satisfy $(\bigwedge G\theta\leftarrow \bigwedge H\theta)\sigma$
  (equivalently, $\forall(\bigwedge G\theta\leftarrow \bigwedge H\theta)$ is 
  skeptically entailed by $P$).
\end{theorem}

However, skeptical resolution is not always complete.  Completeness analysis is founded on ground skeptical derivations, that require a ground version of \CounterSupp.

\begin{definition}
  For all ground atoms $A$, let $\CounterSupp^g(A)$
  be the least set such that if $\langle K,\theta\rangle\in\CounterSupp(A')$
  and for some grounding $\sigma$, $A=A'\theta\sigma$, then 
  \[\langle K\sigma,\epsilon\rangle\in \CounterSupp^g(A),\] 
  where $\epsilon$ is the empty substitution. 
\end{definition}

\begin{theorem}[Finite Ground Completeness, \cite{DBLP:journals/jar/Bonatti01}]
  If some ground implication $\bigwedge G\leftarrow \bigwedge H$
  is skeptically entailed by a \emph{finite} ground program $P$ and 
  $\CounterSupp$ is complete with respect to\ $P$, then $(G \mid H)$ has a successful 
  skeptical derivation from $P$ and $\CounterSupp^g$ with restart 
  goal $G$. In particular, if $G$ is skeptically entailed by 
  $P$, then $(G \mid \emptyset)$ has such a derivation.
\end{theorem}

This basic theorem and the following standard lifting lemma
allow to prove completeness for all finitely recursive normal programs. 

\begin{lemma}[Lifting, \cite{DBLP:journals/jar/Bonatti01}]
  Let $\CounterSupp$ be complete.
  For all skeptical derivations $\mathcal{D}$ from a normal program $P$
  and $\CounterSupp^g$ with restart goal $G_0$,
  there exists a substitution $\sigma$ and a skeptical derivation
  $\mathcal{D'}$ from $P$ and $\CounterSupp$ with restart goal 
  $G_0'$ and answer substitution $\theta$, such that 
  $\mathcal{D}=\mathcal{D'}\theta\sigma$ and $G_0=G_0'\theta\sigma$.
\end{lemma}

\begin{theorem}[Completeness for finitely recursive normal programs]
\label{SkepResComplete}
  Let $P$ be a finitely recursive normal program.
  Suppose $\CounterSupp$ is complete with respect to\ $P$ and that 
  for some grounding substitution $\gamma$, 
  $(\bigwedge G\leftarrow \bigwedge H)\gamma$ holds 
  in all the stable models of $P$. Then $(G \mid H)$
  has a successful skeptical derivation from $P$ 
  and $\CounterSupp$ with restart goal $G$ and some 
  answer substitution $\theta$ more general than $\gamma$.
\end{theorem}
\begin{proof}
  By Theorems \ref{newDefFinitelyRec} and 
  \ref{finiteSkepInf}, there exists
  a smooth module sequence for $P$ with finite elements $P_1$, $P_2$, ..., 
  and a finite $k$ such that 
  $(\bigwedge G\leftarrow \bigwedge H)\gamma$ holds 
  in all the stable models of $P_k$. 
  Since each $P_i$ is downward closed, the ground supports of any given $A \in \atom(P_k)$ with respect to\ program $P_k$ coincide with the ground supports of $A$ with respect to\ the entire program $P$.  Consequently, also ground counter-supports and (generalized) counter-supports, respectively, coincide in $P_k$ and $P$.  Therefore, \CounterSupp is complete with respect to\ $P_k$, too.
  As a consequence, since $P_k$ is a ground, 
  finite program, the ground completeness theorem 
  \hide{(\cite[Theorem~3.17]{DBLP:journals/jar/Bonatti01})}
  can be applied to conclude that
  $(G \mid H)\gamma$ has a successful skeptical derivation from  
  $P_k$ and $\CounterSupp^g$ with restart goal $G\gamma$.  The same derivation is also a derivation from $P$ (as $P_k\subseteq \ground(P)$) and $\CounterSupp^g$.
  Then, by the Lifting lemma (note that $P$ is supposed to be normal), $(G \mid H)$ has a successful 
  skeptical derivation from $P$ and $\CounterSupp$, with 
  restart goal $G$ and some answer substitution $\theta$,
  such that $(G \mid H)\gamma$ is an instance of $(G \mid H)\theta$.
  It follows that $\theta$ is more general than $\gamma$.      
\end{proof}

An important question is whether any computable, complete function \CounterSupp exists.  Take any module sequence $P_1, P_2, \ldots, P_i,\ldots$ based on any effective enumeration of $GH$.  Note that for all ground atoms $A$ one can effectively find a $k\in \mathbb{N}$ such that $A\in\atom(P_k)$.  Now, if the given program $P$ is finitely recursive, then the ground supports of $A$ can be computed by building all the acyclic SLD-derivations for $A$ using the finitely many ground rules of $P_k$.  Consequently,
the ground counter-supports of $A$ are finite and finitely many, too, and can be easily computed from the ground supports of $A$.  Now consider a nonground atom $A$.  Let $\CounterSupp(A)$ be the set of pairs $\tup{K,\gamma}$ such that $\gamma$ is a grounding substitution for $A$ and $K$ is a ground counter-support for $A\gamma$.  Clearly, for any given $A$ such pairs can be recursively enumerated by enumerating the ground instances of $A$, and computing for each of them the corresponding ground counter-supports as explained above.  Clearly, this counter-support function is complete by construction.  This proves that:

\begin{theorem}
If $P$ is normal and finitely recursive, then there exists a complete \CounterSupp function such that for all atoms $A$, $\CounterSupp(A)$ is recursively enumerable.
\end{theorem}

This property allows to recursively enumerate all skeptical derivations from $P$.  Therefore, skeptical resolution provides an alternative proof that skeptical inference from finitely recursive normal programs is in r.e.

\section{Finitary programs and other decidable fragments}
\label{sec:finitary}

The inherent complexity of finitely recursive programs calls for further
restrictions to make deduction decidable.

One of such additional restrictions is based on the following idea:
Suppose that there exists a module sequence $P_1,P_2,\ldots,P_i,\ldots$
and an index $k$ such that for all interpretations $I\subseteq
\atom(P_k)$, the ``top'' program $e_{\atom(P_k)}(\ground(P)\setminus
P_k,I)$ is consistent.  Then the splitting theorem guarantees that
every stable model of $P_k$ can be extended to a stable model of $P$
and, conversely, every stable model of $P$ extends a stable model of $P_k$.
As a consequence, given a ground goal $G$ (be it credulous or
skeptical) whose atoms are included in $\atom(P_k)$, the answer to $G$
can be computed by inspecting only the stable models
$M_{k,1},\ldots,M_{k,n}$ of $P_k$ (which is a finite ground program if
$P$ is finitely recursive). The ``upper'' part of the stable models of
$P$, that is, the stable models of $e_\atom(P_k)(\ground(P)\setminus
P_k,M_{k,i})$ ($1\leq i\leq n$), need not be computed at all---we only
need to know that they exist to be confident that
$M_{k,1},\ldots,M_{k,n}$ are sufficient to answer $G$.

This is the idea underlying \emph{finitary programs}
\cite{DBLP:journals/ai/Bonatti04}.  For normal programs, the
consistency of the top program is guaranteed by means of a theorem due
to Fages \cite{Fag94}, stating that \emph{order consistent} normal
programs are always consistent.  A normal program is order consistent
if there exists no infinite sequence of (possibly repeated) atoms
$\tup{A_i}_{i<\omega}$ such that $A_i$ depends both positively and
negatively on $A_{i+1}$ for all $i<\omega$.  For example, all positive
programs are trivially order consistent, while Fage's program

  \begin{align*}
    q(X)&\leftarrow q(f(X))\\
    q(X)&\leftarrow \naf q(f(X))\\
  \end{align*}
exploited in Example~\ref{counterexa} is not, as well as any normal program whose
dependency graph contains some odd-cycle.  The above program shows
that a program may fail to be order consistent even if the program is
acyclic.  However, if $P$ is normal and
finitely recursive, then it can be shown that $P$ is order consistent
iff $P$ is odd-cycle free \cite{DBLP:journals/ai/Bonatti04}. This
observation justifies the definition of finitary programs
(Definition~\ref{def4}): By requiring finitary programs to have finitely
many odd-cycles, it is possible to confine all odd-cycles into a
single, finite program module $P_k$ and ensure that the ``top''
programs are odd-cycle free and hence consistent.

As proved in \cite{DBLP:journals/ai/Bonatti04}, the extra condition on
odd-cycles suffices to make both credulous and skeptical ground
queries decidable.  However, in \cite{DBLP:journals/ai/Bonatti04} the
statement erroneously fails to include the set of odd-cyclic literals
among the inputs of the algorithm.  Here is the correct statement and
a slightly different proof based on module sequences:

\begin{theorem}
  Given a finitary normal program $P$ and a finite set $C$ containing (at
  least) all of the odd-cyclic
  atoms of $P$'s Herbrand base,
  \begin{enumroman}
  \item deciding whether a ground formula $G$ is a credulous
    consequence of $P$ is decidable;
  \item deciding whether a ground formula $G$ is a skeptical
    consequence of $P$ is decidable.
  \end{enumroman}
\end{theorem}

\begin{proof} (Sketch)
  Let $P_1,P_2,\ldots,P_i,\ldots$ be any (recursive) module sequence
  induced by a recursive enumeration of $P$'s Herbrand base, and let
  $k$ be the minimal index such that $C \cup \atom(G) \subseteq
  \atom(P_k)$. Clearly, such a $k$ exists and is effectively
  computable.  Moreover, $P_k$ is ground and finite (because $P$ is
  finitely recursive), therefore the set of its stable models
  $M_{k,1},\ldots,M_{k,n}$ can be effectively computed as well, it is
  finite, and consists of finite models.  Now, by construction, the
  ``top'' programs $e_{\atom(P_k)}(\ground(P)\setminus P_k,M_{k,i})$
  ($1\leq i\leq n$) are all odd-cycle free---and hence consistent, by
  Fage's theorem.  It follows by the splitting theorem that for all
  $i=1,\ldots,n$, the program $P$ has a stable model $M$ such that $M\cap
  \atom(P_k)=M_{k,i}$.  As a consequence, if $G$ is true (resp.\
  false) in a stable model of $P_k$, then $G$ must be true (resp.\
  false) in a stable model of $P$.  Conversely, by the splitting
  theorem, if $G$ is true (resp.\ false) in a stable model of $P$,
  then $G$ must be true (resp.\ false) in a stable model of $P_k$
  (because $\atom(P_k)$ splits $P$).  It follows easily that $G$ is a
  credulous (resp.\ skeptical) consequence of $P$ iff $G$ is a
  credulous (resp.\ skeptical) consequence of $P_k$.  Of course, since
  the set of stable models of $P_k$ is finite, recursive, and contains
  only finite models, both the credulous and the skeptical consequences of
  $P_k$ are decidable.  
\end{proof}

Extending this result to disjunctive programs is not a trivial
task because, unfortunately, Fage's theorem does not scale to disjunctive
programs in any obvious way.
Consider the possible natural generalization of atom dependencies from the class of normal programs to the class of disjunctive programs:

\begin{enumerate}
\item First assume that the unlabelled edges of $DG(P)$ are ignored,
  that is, let $A$ depend on $B$ iff there is a path from $A$ to $B$
  in $DG(P)$ with no unlabelled edges.
  This is equivalent to adopting a dependency graph similar to the
  traditional graphs for normal programs, with no head-to-head edges.
  Using the resulting notion of atom dependencies, one can find
  programs that are order consistent but have no stable models.  One
  of them is
  \begin{eqnarray*}
    & p_1 \lor p_2 \\
    & q_1 \lor q_2 \\
    & p_1 \leftarrow \naf q_1\\
    & q_1 \leftarrow \naf p_2\\
    & p_2 \leftarrow \naf q_2\\
    & q_2 \leftarrow \naf p_1 \,.
  \end{eqnarray*}

\item Next, suppose that unlabelled edges are regarded as positive
  edges, that is, $A$ depends positively (resp.\ negatively) on $B$
  iff there is a path from $A$ to $B$ in $DG(P)$ with an even (resp.\
  odd) number of negative edges.  The above inconsistent program
  is still order consistent under this new notion of dependency.

\item Finally, assume that unlabelled edges are regarded as negative edges.
  This is a natural assumption given the minimization-based nature of
  disjunctive stable models: For instance if $P=\{p\lor q\}$, then the
  falsity of $p$ implies the truth of $q$ and viceversa (indeed $P$ is
  equivalent to $\{p\leftarrow \naf q,\ q\leftarrow \naf p\}$).  
  A major problem is that with this form of
  dependency, too many interesting
  disjunctive programs are \emph{not} order consistent:
  \begin{itemize}
  \item every rule with at least three atoms in the head generates an
    odd-cycle through those atoms, therefore the program would not be
    order consistent;
  \item for every cycle $\cal C$ containing a head-to-body edge $(A,\pm,B)$
    originated by a ``proper'' disjunctive rule (i.e., a rule with two
    or more atoms in the head) there exists an odd-cycle
    (possibly $\cal C$ itself, or the cycle obtained by extending
    $\cal C$ with a
    negative edge from $A$ to another atom in the same head).  This
    means that disjunctive rules could never be applied in any recursion.
  \end{itemize}
\end{enumerate}

Similar problems (preconditions that are difficult to ensure in
practical cases) affect Turner's approach to consistency
\cite{DBLP:conf/slp/Turner94}.  His \emph{signed programs} generalize
order consistent normal programs as follows: It should be possible to partition
the Herbrand base into two sets $H_1$ and $H_2$ such that:
\begin{enumerate}
\item negative edges always cross the two partitions;  positive edges
  never do;
\item each rule head is entirely contained in a single partition;
\item the set of rules whose
  head is contained in $H_1$ is a normal program.
\end{enumerate}
Unfortunately, to the best of our knowledge no application domains
naturally require programs satisfying the third condition (that
roughly speaking makes the program ``half normal'').

A more recent paper \cite{DBLP:conf/iclp/Bonatti02} ensures
consistency through the theory of program \emph{shifting}
\cite{Bo-shifts}.  A shifting of $P$ is a modified version of $P$
where some atoms are moved from heads to bodies and enclosed in the
scope of a negation symbol. This transformation preserves the
classical semantics of the program but not its stable models.
However, every stable model of a shifted program is also a stable
model of the original program, so the consistency of the former
implies the consistency of the latter.  Then the approach of
\cite{DBLP:conf/iclp/Bonatti02} consists in adding more conditions to
the definition of finitary programs to ensure that at least one
``full'' shifting of $P$---transforming $P$ into a normal program---is
finitary, so that the original consistency theorem by Fages can be
applied.  The main drawback of this approach is that the extra conditions
required are clumsy and---again---difficult to use in practice.

A very interesting and novel recent approach by Eiter and Simkus
\cite{DBLP:conf/lpar/SimkusE07} consists in replacing the consistency
property with other properties enjoyed by some decidable fragments of
first-order logic such as description logics and the guarded fragment.
In these fragments, consistent theories always have both a finite
model and a tree model which is the ``unwinding'' of the finite model,
i.e., a regular tree.  Syntactic restrictions on predicate arity and
on the occurrences of function symbols (modelled around the
skolemization of guarded formulae) have been exploited to prove the
decidability of a new class of finitely recursive programs called FDNC
programs.  In our framework, this idea roughly corresponds to having
regular module sequences where after some steps the new rules
contained in $P_i\setminus P_{i-1}$ are isomorphic to some previous
program slice $P_j\setminus P_{j-1}$ ($j<i$).  Therefore in order to
find a stable model of $P$ one needs only to find a stable model $M$
for some finite module $P_i$, as a model for the upper part can then
be constructed by cloning $M$ or submodels thereof.  FDNC
programs can be applied to encode ontologies expressed in description
logics, and are suitable to model a wide class of planning problems.
An interesting open question is whether this approach can be
generalized to wider interesting classes of programs by studying
regular module sequences.

\section{Conclusions}
\label{conclusions}

In this paper we have extensively studied the properties of stable
model reasoning with disjunctive, finitely recursive programs---a very
expressive formalism for answer set programming.
Finitely recursive programs extend the class of finitary 
programs by dropping the restrictions on odd-cycles, that is, on the number of possible sources of inconsistencies.
We extended to finitely recursive programs many of the nice properties
of finitary programs: (i) a compactness property (Theorem~\ref{Pcompactness}); (ii) the
r.e.-completeness of inconsistency checking and skeptical 
inference (Theorem~\ref{InconsPreComplete}); (iii)
the completeness of skeptical resolution
(Theorem~\ref{SkepResComplete}); note that this result applies to
normal programs only, unlike (i) and (ii).

Unfortunately, some of the nice properties of finitary programs do \emph{not}
carry over to finitely recursive programs: (i) ground queries are not
decidable (Theorem~\ref{th:skep-r-e-complete} and Corollary~\ref{credInf-co-r-e}); 
(ii) nonground credulous
queries are not semidecidable (Corollary~\ref{credInf-co-r-e}).  

We proved our results by extending the splitting sequence theorem
that, in general, guarantees only that each consistent
program $P$ has a consistent module sequence
for $P$.  We proved that in general the converse does not hold
(Example~\ref{counterexa}), unless $P$ is finitely recursive: In that
case, the stable models of a consistent module sequence always
converge to a model of $P$ (Theorem~\ref{ConsSplitSeqPCons}).

As a side benefit, our techniques introduce a normal form for
splitting sequences and their bottom programs, where sequence length is limited to
$\omega$ and---if the program is finitely recursive---the sequence is
smooth (i.e., the ``delta'' between each non-limit element and its
predecessor is finite).  Such properties constitute an alternative
characterization of finitely recursive programs.  The theory of module
sequences is a powerful tool for working on decidable inference with
infinite stable models, as it provides a constructive, iterative
characterization of the stable models of a large class of programs
with infinite domains.  In Section~\ref{sec:finitary} we carried out a
first attempt at relating different approaches using module sequences
as a unifying framework.  However such an analysis is still very preliminary
and partially informal; its development constitutes an interesting
subject for future work, and it may contribute to recent areas such as
research on FDNC programs.

Another interesting open problem is extending to disjunctive programs
Fage's consistency result (an important ingredient in several
decidability results).  The existing approaches are based on rather
restrictive assumptions that call for more flexible solutions.

Finally, an interesting theoretical question is whether skeptical
resolution can be extended to disjunctive programs. A related
challenge is finding a satisfactory goal-directed calculus for the positive
fragment, which is based on a minimal model semantics.

\subsection*{Acknowledgements}

This work was partially supported by the PRIN project \emph{Enhancement and Applications of Disjunctive Logic Programming}, funded by the Italian Ministry of Research (MIUR).

\bibliography{bib}

\begin{thebibliography}{}

\bibitem[\protect\citeauthoryear{Baral}{Baral}{2003}]{Baral03}
{\sc Baral, C.} 2003.
\newblock {\em {Knowledge Representation, Reasoning and Declarative Problem
  Solving}}.
\newblock Cambridge University Press, Cambridge.

\bibitem[\protect\citeauthoryear{Baselice, Bonatti, and Criscuolo}{Baselice
  et~al\mbox{.}}{2007}]{DBLP:conf/iclp/BaseliceBC07}
{\sc Baselice, S.}, {\sc Bonatti, P.~A.}, {\sc and} {\sc Criscuolo, G.} 2007.
\newblock On finitely recursive programs.
\newblock In {\em ICLP}, {V.~Dahl} {and} {I.~Niemel{\"a}}, Eds. Lecture Notes
  in Computer Science, vol. 4670. Springer, 89--103.

\bibitem[\protect\citeauthoryear{Bonatti}{Bonatti}{2001a}]{BoLPNMR01}
{\sc Bonatti, P.} 2001a.
\newblock Prototypes for reasoning with infinite stable models and function
  symbols.
\newblock In {\em Logic Programming and Nonmonotonic Reasoning, 6th
  International Conference, LPNMR 2001}. LNCS, vol. 2173. Springer, 416--419.

\bibitem[\protect\citeauthoryear{Bonatti}{Bonatti}{1993}]{Bo-shifts}
{\sc Bonatti, P.~A.} 1993.
\newblock Shift-based semantics: {G}eneral results and applications.
\newblock Tech. Rep. CD-TR 93/59, Technical University of Vienna, Computer
  Science Department, Institute of Information Systems.

\bibitem[\protect\citeauthoryear{Bonatti}{Bonatti}{2001b}]{DBLP:journals/jar/B%
onatti01}
{\sc Bonatti, P.~A.} 2001b.
\newblock Resolution for skeptical stable model semantics.
\newblock {\em J. Autom. Reasoning\/}~{\em 27,\/}~4, 391--421.

\bibitem[\protect\citeauthoryear{Bonatti}{Bonatti}{2002}]{DBLP:conf/iclp/Bonat%
ti02}
{\sc Bonatti, P.~A.} 2002.
\newblock Reasoning with infinite stable models {II}: Disjunctive programs.
\newblock In {\em ICLP}, {P.~J. Stuckey}, Ed. Lecture Notes in Computer
  Science, vol. 2401. Springer, 333--346.

\bibitem[\protect\citeauthoryear{Bonatti}{Bonatti}{2004}]{DBLP:journals/ai/Bon%
atti04}
{\sc Bonatti, P.~A.} 2004.
\newblock Reasoning with infinite stable models.
\newblock {\em Artif. Intell.\/}~{\em 156,\/}~1, 75--111.

\bibitem[\protect\citeauthoryear{Calimeri, Cozza, Ianni, and Leone}{Calimeri
  et~al\mbox{.}}{2008}]{DBLP:conf/iclp/CalimeriCIL08}
{\sc Calimeri, F.}, {\sc Cozza, S.}, {\sc Ianni, G.}, {\sc and} {\sc Leone, N.}
  2008.
\newblock Computable functions in {ASP}: {T}heory and implementation.
\newblock In {\em ICLP}, {M.~G. de~la Banda} {and} {E.~Pontelli}, Eds. Lecture
  Notes in Computer Science, vol. 5366. Springer, 407--424.

\bibitem[\protect\citeauthoryear{Eiter, Leone, Mateis, Pfeifer, and
  Scarcello}{Eiter et~al\mbox{.}}{1997}]{DLV}
{\sc Eiter, T.}, {\sc Leone, N.}, {\sc Mateis, C.}, {\sc Pfeifer, G.}, {\sc
  and} {\sc Scarcello, F.} 1997.
\newblock A deductive system for non-monotonic reasoning.
\newblock In {\em Logic Programming and Nonmonotonic Reasoning, 4th
  International Conference, LPNMR'97, Proceedings}. LNCS, vol. 1265. Springer,
  364--375.

\bibitem[\protect\citeauthoryear{Fages}{Fages}{1994}]{Fag94}
{\sc Fages, F.} 1994.
\newblock {C}onsistency of {C}lark's completion and existence of stable models.
\newblock {\em Methods of Logic in Computer Science\/}~{\em 1}, 51--60.

\bibitem[\protect\citeauthoryear{Gelfond and Lifschitz}{Gelfond and
  Lifschitz}{1988}]{GL88}
{\sc Gelfond, M.} {\sc and} {\sc Lifschitz, V.} 1988.
\newblock The stable model semantics for logic programming.
\newblock In {\em Proc. of the 5th ICLP}. MIT Press, 1070--1080.

\bibitem[\protect\citeauthoryear{Gelfond and Lifschitz}{Gelfond and
  Lifschitz}{1991}]{gelfond91classical}
{\sc Gelfond, M.} {\sc and} {\sc Lifschitz, V.} 1991.
\newblock Classical negation in logic programs and disjunctive databases.
\newblock {\em New Generation Computing\/}~{\em 9,\/}~3-4, 365--386.

\bibitem[\protect\citeauthoryear{Lifschitz and Turner}{Lifschitz and
  Turner}{1994}]{lifschitz94splitting}
{\sc Lifschitz, V.} {\sc and} {\sc Turner, H.} 1994.
\newblock Splitting a logic program.
\newblock In {\em International Conference on Logic Programming}. MIT Press,
  23--37.

\bibitem[\protect\citeauthoryear{Lloyd}{Lloyd}{1984}]{DBLP:books/sp/Lloyd84}
{\sc Lloyd, J.~W.} 1984.
\newblock {\em Foundations of Logic Programming, 1st Edition}.
\newblock Springer.

\bibitem[\protect\citeauthoryear{Marek and Remmel}{Marek and
  Remmel}{2001}]{MR01expressibility}
{\sc Marek, V.} {\sc and} {\sc Remmel, J.} 2001.
\newblock On the expressibility of stable logic programming.
\newblock In {\em Logic Programming and Nonmonotonic Reasoning, 6th
  International Conference, LPNMR 2001}. LNCS, vol. 2173. Springer, 107--120.

\bibitem[\protect\citeauthoryear{Marek and Truszczynski}{Marek and
  Truszczynski}{1998}]{DBLP:journals/corr/cs-LO-9809032}
{\sc Marek, V.~W.} {\sc and} {\sc Truszczynski, M.} 1998.
\newblock Stable models and an alternative logic programming paradigm.
\newblock {\em CoRR\/}~{\em cs.LO/9809032}.

\bibitem[\protect\citeauthoryear{Niemel{\"a}}{Niemel{\"a}}{1999}]{DBLP:journal%
s/amai/Niemela99}
{\sc Niemel{\"a}, I.} 1999.
\newblock Logic programs with stable model semantics as a constraint
  programming paradigm.
\newblock {\em Ann. Math. Artif. Intell.\/}~{\em 25,\/}~3-4, 241--273.

\bibitem[\protect\citeauthoryear{Niemel{\"a} and Simons}{Niemel{\"a} and
  Simons}{1997}]{smodels}
{\sc Niemel{\"a}, I.} {\sc and} {\sc Simons, P.} 1997.
\newblock Smodels -- an implementation of the stable model and well-founded
  semantics for normal {LP}.
\newblock In {\em Logic Programming and Nonmonotonic Reasoning, 4th
  International Conference, LPNMR'97, Proceedings}. LNCS, vol. 1265. Springer,
  421--430.

\bibitem[\protect\citeauthoryear{Simkus and Eiter}{Simkus and
  Eiter}{2007}]{DBLP:conf/lpar/SimkusE07}
{\sc Simkus, M.} {\sc and} {\sc Eiter, T.} 2007.
\newblock {FDNC}: Decidable non-monotonic disjunctive logic programs with
  function symbols.
\newblock In {\em 14th Int. Conf. on Logic for Programming, Artificial
  Intelligence, and Reasoning, LPAR 2007}. Lecture Notes in Computer Science,
  vol. 4790. Springer, 514--530.

\bibitem[\protect\citeauthoryear{Turner}{Turner}{1994}]{DBLP:conf/slp/Turner94}
{\sc Turner, H.} 1994.
\newblock Signed logic programs.
\newblock In {\em SLP}. MIT Press, 61--75.

\bibitem[\protect\citeauthoryear{Turner}{Turner}{1996}]{turner96splitting}
{\sc Turner, H.} 1996.
\newblock Splitting a default theory.
\newblock In {\em Proceedings of the Thirteenth National Conference on
  Artificial Intelligence}, {H.~Shrobe} {and} {T.~Senator}, Eds. AAAI Press,
  Menlo Park, California, 645--651.

\end{thebibliography}
\bibliographystyle{acmtrans}

\end{document}